\documentclass[manuscript,screen,nonacm]{acmart}

\AtBeginDocument{%
  \providecommand\BibTeX{{%
    \normalfont B\kern-0.5em{\scshape i\kern-0.25em b}\kern-0.8em\TeX}}}

\copyrightyear{2022}

\begin{document}

\title{Feng-Shui Compass: A Modern Exploration of Traditional Chinese Environmental Analysis }

\author{Xuanyu “Xy” Fang}
\authornote{All the authors contributed equally to this research.}
\email{xf48@cornell.edu}
\author{Yunzhu Pan}
\authornotemark[1]
\email{yp432@cornell.edu}
\author{Hongjun Wu}
\authornotemark[1]
\email{hw434@cornell.edu}
\affiliation{%
  \institution{Cornell Tech}
  \streetaddress{2 W Loop Rd.}
  \city{New York}
  \state{NY}
  \country{United States}
  \postcode{10044}
}

\renewcommand{\shortauthors}{Fang, Pan and Wu}

\begin{abstract}
  The technological advancement in data analysis and sensor technology has contributed to a growth in knowledge of the surrounding environments. Feng Shui, the Chinese philosophy of evaluating a certain environment and how it influences human well-being, can only be determined by self-claimed specialists for the past thousands of years. We developed a device as well as a procedure to evaluate the ambient environment of a room to perform a study that attempts to use sensor data to predict a person’s well-being score in that environment, therefore evaluating the primary aspect of Feng Shui. Our study revealed preliminary results showing great potential for further research with larger experiments.
\end{abstract}



\maketitle

\section{Introduction}
Feng Shui has been a significant ancient practice of Chinese culture for a long time. It was known and adopted in many forms, while the most important role of it remains a posteriori analysis of the environment to guide people through their everyday lives. This research adopted the traditional practices with scientific methods aiming to explore the possibility of quantifying Feng Shui in a standardized way using modern sensors and technological approaches like machine learning. People’s subjective well-being has been measured in this research as the prediction target of the environmental information in that specific room. The preliminary results suggest that this approach has the potential to assist and explain the Feng Shui rituals in certain aspects and can be further investigated.

\section{RELATED WORK}

Feng Shui originated from ancient people’s empirical recognition of the environment. It guided ancient Chinese for thousands of years on how to perform activities, how to design houses, and how to please ancestors. The modern Feng Shui practices fall into an interesting gap between skeptical ghost-hunting and empirical environmental science. Despite its limits in empirical evidence and consistent results, Feng Shui was still proven to be meaningful in numerous fields: architecture, urban planning, sociology, etc. (LEE, 1986).

One of the most related fields is modern Architecture. Previous research suggests that the formed principles of Feng Shui were based on the fundamental understanding of the physical configuration of geographical features. The many models proposed by the Feng Shui concepts are mostly imaginary ways to explain the basic form of this world in micro spaces. Empirical evidence suggests that the Feng Shui theory was concurred by modern architects in terms of the selection of surrounding environments and interior layout \cite{Mak05}. 
Environmental psychology is also a field that was commonly associated with Feng Shui beyond architecture \cite{Bonaiuto10}. Environmental psychology studied the relationship between people and socio-physical features of the environment. It systematically explains the complex interactions between people and the environment around them in either long-term or short-term effects. Objective methods were used to study the connection between Feng Shui principles and environmental psychology, and the discipline shows a common trend of exploring the subjective well-being of people presenting in a specific environment. 

In order to study the well-being of people in a certain space with scientific methods and systematic approaches, the environmental factor can be divided into various specific measurements. Take air quality as an example, it has been proven to be significant in predicting the well-being of people across situations. Well-controlled air quality is able to improve learning activities in libraries \cite{Righi02} and living experiences within the household \cite{Barrington19}. Air quality consists of a number of significant gas readings that people can be easily exposed to in everyday life: dust, polluting and toxic compounds (benzene, xylene, and toluene), H2, CO2, etc… 

Other ambient factors like temperature and noise level have been more obvious and tangible to humans. Despite their visibility and people’s awareness, they remain a significant influencer of people’s emotional and physical well-being. Heat exposure predicts lower positive emotions, and a higher chance of feeling fatigue \cite{Noelke16}.  Controlling noise levels was valued in urban residential designs, but modern in-house machines like washers, dryers, and air conditioning systems still have the potential to affect people’s living experiences. The noise level was associated with people’s ability to perceive the availability of nearby natural environments, which consecutively influenced people’s ability to maintain a stable emotional status \cite{Gidlof07}. 

Aside from all the micro aspects of environmental factors, there are also macro scopes being focused by Feng Shui in predicting people’s well-being in certain spaces. A research patent published a few years ago successfully developed a mathematical computer vision model to predict a Feng Shui score by recognizing features from floor plans of residential households following traditional Feng Shui principles \cite{Wang14}. The model includes the calculated ratio between the width and length of the room, the facing direction of the window, and the overall space.

\section{RESEARCH QUESTION \& HYPOTHESIS}

Based on previous research, our main research interest lies in developing a modern technological system that will be able to assist and possibly standardize the Feng Shui rituals. The main focus of the Feng Shui rituals and practices was limited to the scope of environmental analysis of certain living spaces. The main hypothesis of this research is that we can use environmental information like sensor data and floor plans to predict people’s subjective well-being when present in that certain space. 

\section{DATA COLLECTION}

\subsection{Environmental Data}

It is very crucial to understand the size and composition of the environment in order to perform a good analysis of the overall Feng Shui of such an environment. In order to achieve this, we collected the width and height of the room, in feet from the participant’s environment.

\subsubsection{Arduino Oplá}

\begin{figure}[h]
  \centering
  \includegraphics[width=\linewidth]{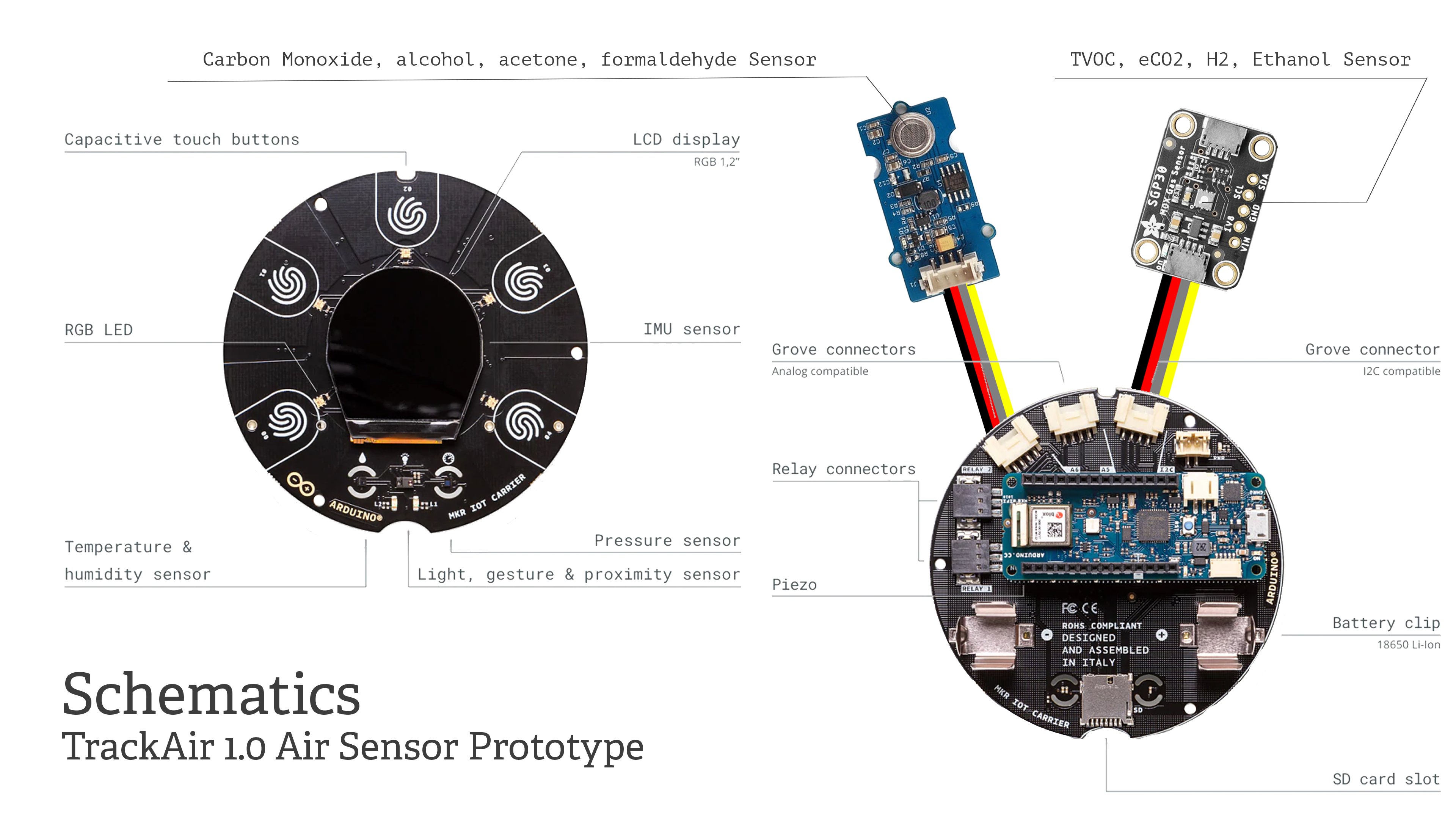}
  \caption{Arduino Oplá Data Collector Schematics}
  \label{fig:opla_sch}
  \Description{Schematics of the Arduino Oplá Data Collector}
\end{figure}

\begin{figure}[h]
  \centering
  \includegraphics[width=0.4\linewidth]{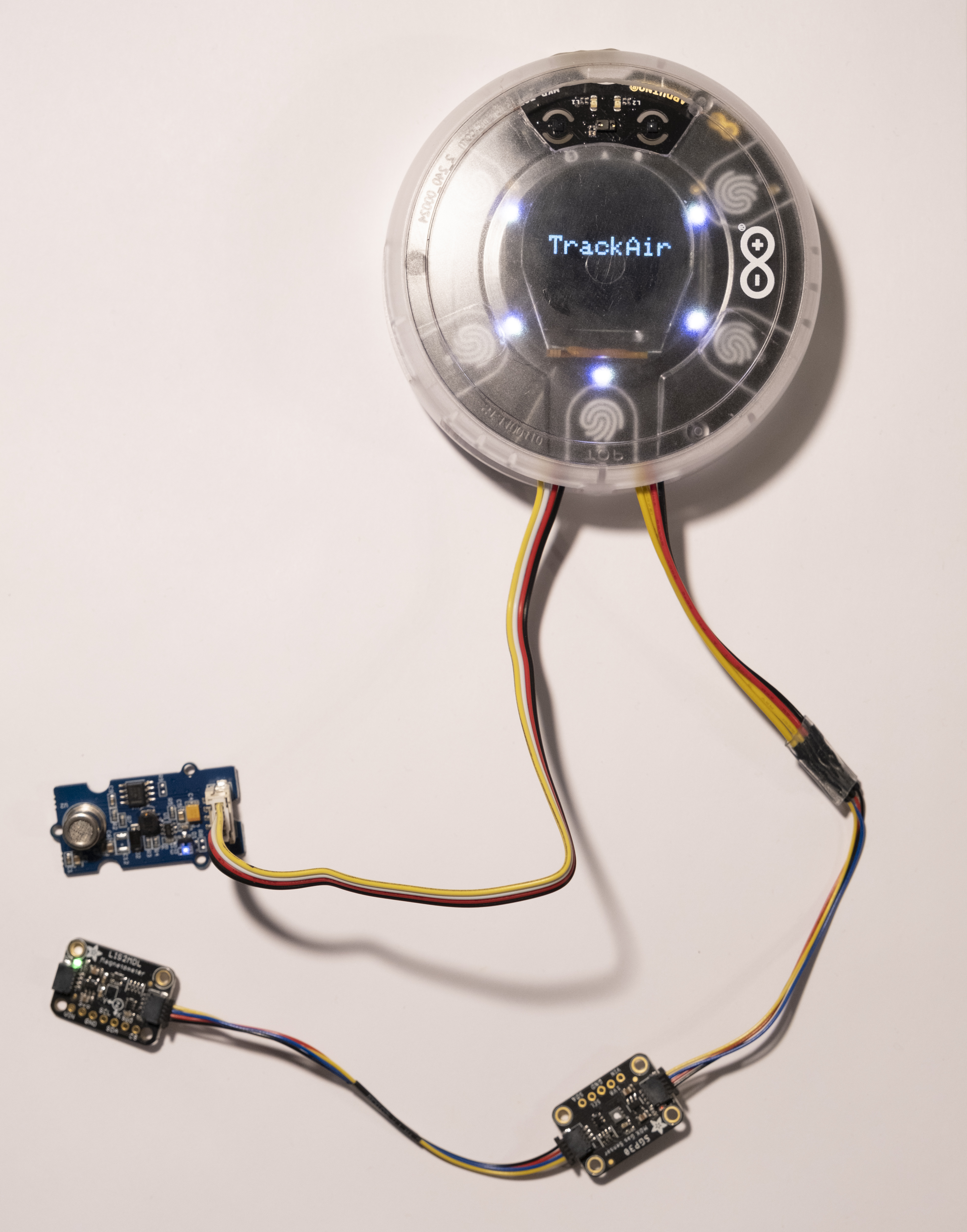}
  \caption{Arduino Oplá Data Collector Photo}
  \label{fig:opla_photo}
  \Description{Photo of the Arduino Oplá Data Collector}
\end{figure}

A data acquisition device developed using Arduino Oplá played a central role in terms of data acquisition. Schematics of the data acquisition device is shown in Figure \ref{fig:opla_sch}. It is then named $TrackAir 1.0$. Figure \ref{fig:opla_photo} is a photo of the deployed device.

There are many factors in the environment that will affect a person’s well-being. We collected all the data that we can collect using the Arduino Oplá as well as other attached air sensors, including temperature, humidity, air pressure, light intensity, toxic chemical level, TVOC level, eCO2 level, H2 level, and Ethanol level. This instrument will sample all the above data twice every second, and it will take 1,000 samples for each participant, which is roughly ten minutes of continuous data collection. Doing so will help us eliminate outliers in the data, which is often caused by a spike in the sensor and is common among consumer-level sensors. We believe these data would be a very comprehensive representation of the environment that our participants are in.

\subsubsection{Smart Phone \& Smart Watch (iPhone \& Apple Watch)}

We collected data about the participant’s room orientation using an iPhone. Using the “Compass” built-in app in iOS, we measured the door direction and desk direction of the room in degrees, as well as whether the room is a rectangle. In order to collect the current state of the participant, we used an Apple Watch to record the heart rate of the participant. Apple Watch was also used to measure the ambient noise in the current environment, which also plays an important role in terms of evaluating a person’s well-being in a given room.

\subsection{Self-Reported Well-being Score: A Survey}

The survey takes around 15 minutes for participants to complete. Participants are required to complete this survey in person in the same environment where the data is collected.

\subsubsection{Part 1 (1 Minute): Demographics.}

Participants record their basic demographic data.

\subsubsection{Part 2 (5-10 Minutes): Respond to Neutral images from the “GAPED Picture Database”.}

Participants were asked to repeat the following procedure ten times, each time with one different picture.
 Participants were prompted using the following prompt, which was written with a neutral tone in mind.
 Please view the following pictures and be mindful of how you feel within this specific space you are currently sitting at.
Participants were asked to view an image, which is selected from the “GAPED Picture Database” that has been scientifically proven to be “Neutral”.
 Participants were prompted with the following neutral tone prompt.
 Please use one word to describe what's in this image.
Participants were instructed to type one word to describe what is in the image, to ensure they are focused and using critical thinking (System 2 thinking).
Participants were prompted to respond to “How does this image make you feel?” and select a number between 0-5, 0 being the most negative and 5 being the most positive.

\subsubsection{Part 3 (5-10 Minutes): Mini MASQ Well-Being Assessment.}

Participants are required to fill out the MASQ Well-Being Assessment. We used the mini version of the assessment, which consists of 26 questions. The entire questionnaire, as well as scoring methods, were extracted from the Mood and Anxiety Symptom Questionnaire \cite{Boschen07}. 15 questions within this questionnaire were reverse coded to ensure reliability. Using this survey, a wellness score was computed for each participant using the average of the reported wellness, a continuous number between 1 to 5. 1 being the most negative and 5 being the most positive. This score was then used as the ground truth for data analysis.

\section{Experiments \& Results}

\subsection{Feature Engineering}

\begin{table}
  \caption{All Features}
  \label{tab:all_feat}
  \begin{tabular}{ll}
    \toprule
    Smart Phone Features & Arduino Opla Features\\
    \midrule
    Width & Temperature\\
    Height & Humidity\\
    Width-Height Ratio Score & Light Intensity\\
    isRectangle & Air pressure Level\\
    Door Direction & Toxic Chemical Level\\
    Desk Direction & H2 Level\\
    Noise Level & eCO2 Level\\
     & TVOC Level\\
     & Ethanol Level\\
  \bottomrule
 \end{tabular}
\end{table}

Using the Arduino Oplá data collection device, in each room, we collected 1000 entries of all feature data shown in the right column of Table \ref{tab:all_feat} across a time span of roughly 10 minutes, at the same time the participants were filling out the Self-Reported Well-being Survey. We then calculated the mean and standard deviation of each feature across all entries, which are added to the iPhone-collected features, as shown in the left column of Table \ref{tab:all_feat}, to form a complete feature dataset.
It’s worth noting that we adopted the calculation of the BestRatio score of a given room, using formula \ref{eq:best_ratio} as proposed in \cite{Wang14}. The Width-Height ratio is adopted as one of the core features in our dataset.

\begin{equation}
\label{eq:best_ratio}
s_{i}\left\{\begin{array}{c}
\sin \left(\frac{\text { width }}{\text { length }} \times \frac{\pi}{2}\right), \frac{\text { width }}{\text { length }} \leq \text { BestRatio } \\
2 \times \sin \frac{\text { BestRatiox }}{2}-\sin \left(\frac{\text { width }}{\text { length }} \times \frac{\pi}{2}\right), \frac{\text { width }}{\text { length }}>\text { BestRatio }
\end{array}\right.
\end{equation}

\subsection{Data Analysis \& Prediction}

As of the time the report is concluded, we collected 22 pieces of data in total, each piece consisting of 26 environmental features and a well-being score calculated from the survey as described above. The size of the dataset is too small to train any off-the-shelf state-of-the-art classification model. To verify this assumption, we tried training the KNN classifier with all collected features included, and the model resulted in very poor performance. This is due to the fact that the dimension of model parameters from all 26 features is too complicated to be trained on a dataset so small. Thus, we proceeded to analyze the correlation between each feature and the self-reported well-being score, in order to try to shrink the feature dimension. For each feature, we calculated the Pearson product-moment correlation coefficients, which calculate the relationship between the correlation coefficient matrix (R) from the covariance matrix(C) as follows in 
Formula \ref{eq:corr_formula}.  The results are shown in Table \ref{tab:feat_corr}. The values of correlation coefficients are between -1 and 1, inclusive, with a larger absolute value indicating a strong correlation between the feature and the corresponding well-being score.

\begin{equation}
\label{eq:corr_formula}
R_{i j}=\frac{C_{i j}}{\sqrt{C_{i i} * C_{j j}}}
\end{equation}

\begin{table}
  \caption{Correlation Coefficients of Features}
  \label{tab:feat_corr}
  \begin{tabular}{lr}
    \toprule
    Feature Name & Correlation Coefficient\\
    \midrule
    width & -0.24338773691387658\\
    height & -0.29759112270921384\\
    door\_direction & 0.24677463268661320\\
    desk\_direction & -0.11004988036162887\\
    is\_rect & 0.05590169943749470\\
    noise\_db & -0.30224131068722480\\
    Temperature\_mean & 0.03247514240073649\\
    Humidity\_mean & 0.07288029553127462\\
    Air\_Pressure\_mean & -0.14634161342130952\\
    Light\_Intensity\_mean & -0.35336497334816450\\
    Toxic\_Chemical\_Level\_mean & -0.01665129618237642\\
    TVOC\_Level\_mean & -0.34491819179506240\\
    eCO2\_Level\_mean & 0.35916354101294357\\
    H2\_Level\_mean & -0.02958013062252979\\
    Ethanol\_Level\_mean & -0.06445525969167733\\
    Temperature\_std & 0.11654017410369533\\
    Humidity\_std & 0.00941482862018411\\
    Air\_Pressure\_std & 0.04834266586644965\\
    Light\_Intensity\_std & -0.18967663877532690\\
    Toxic\_Chemical\_Level\_std & -0.04258358820208688\\
    TVOC\_Level\_std & -0.01901284691080757\\
    eCO2\_Level\_std & 0.29220914867111870\\
    H2\_Level\_std & 0.30894787538751880\\
    Ethanol\_Level\_std & 0.08888045959606182\\
    wh\_ratio\_score & -0.25847362340400610\\
  \bottomrule
 \end{tabular}
\end{table}

We then selected the features with a correlation score higher than 0.2 as candidates for classifier training. We experimented with all possible combinations of the high-correlation candidates, and the set of candidates shown in Table \ref{tab:best_candids} yielded the best classifier results. 

\begin{table}
  \caption{Beat Performing Feature Set}
  \label{tab:best_candids}
  \begin{tabular}{ll}
    \toprule
    Feature Names &\\
    \midrule
    width & TVOC\_Level\_mean\\
    height & Light\_Intensity\_mean\\
    wh\_ratio & eCO2\_Level\_mean\\
    door\_direction & H2\_Level\_std\\
    noise\_db & eCO2\_Level\_std\\
  \bottomrule
 \end{tabular}
\end{table}

\begin{figure}[h]
  \centering
  \includegraphics[width=0.8\linewidth]{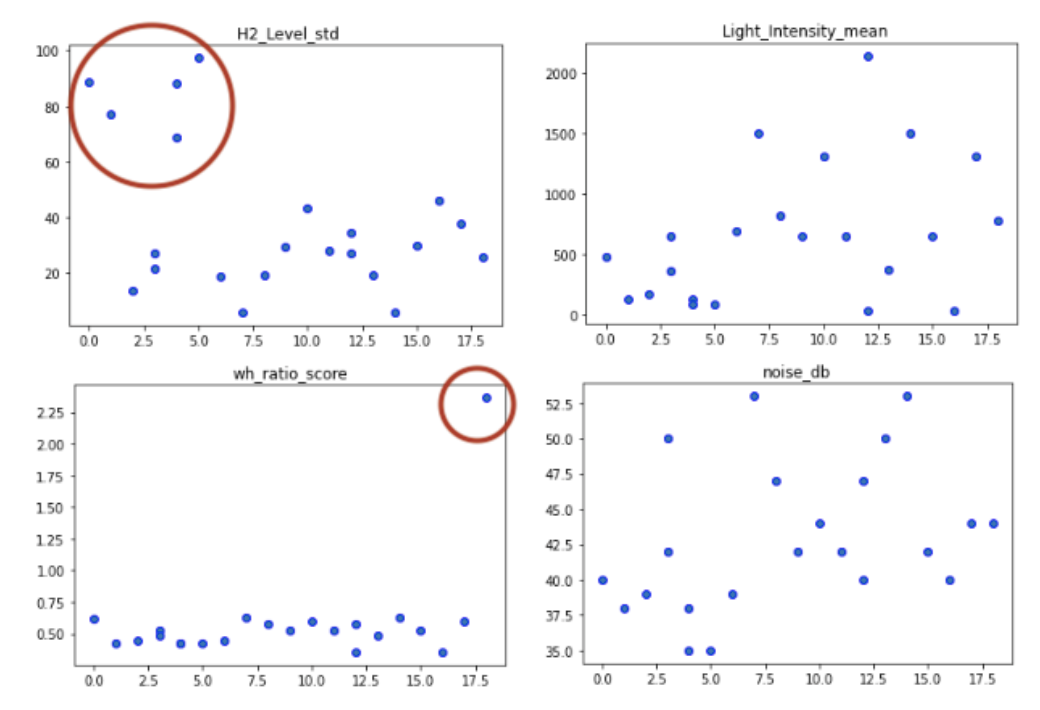}
  \caption{Feature Distribution Examples}
  \Description{Examples of feature's distribution pattern.}
\end{figure}

With a closer look at the data distribution of these features, we find out that the effectiveness of candidates is largely determined by their distribution “cleanness”. For example, as circled out in red circles, features H2\_level\_std and wh\_ratio\_score have outliers. Given the small sample size, outliers have a strong negative impact on model training. On the contrary, good-performing candidates such as Light\_intensity\_mean and noise\_db have no outliers disturbing the data distributions.

\subsection{Model Prediction}

\begin{figure}[h]
  \centering
  \includegraphics[width=0.5\linewidth]{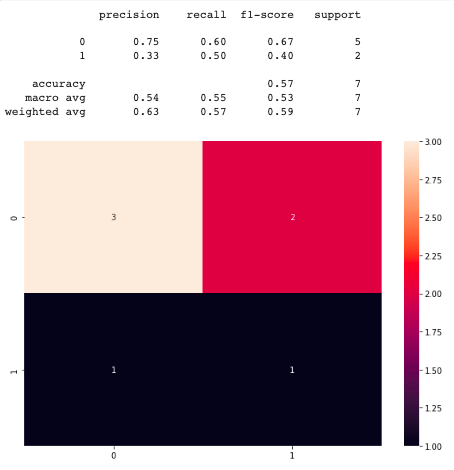}
  \caption{KNN Classifier Results}
  \label{res:knn}
  \Description{Precision, recall, f1 scores and accuracy of KNN Classifier.}
\end{figure}

\begin{figure}[h]
  \centering
  \includegraphics[width=0.5\linewidth]{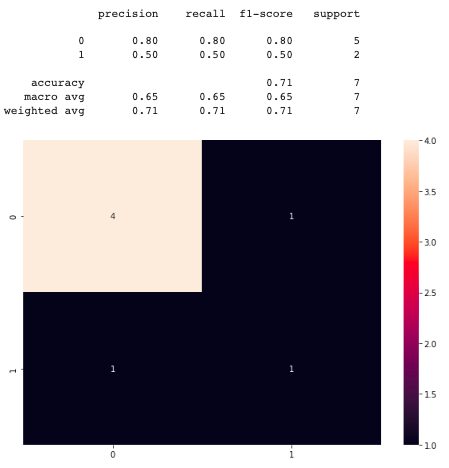}
  \caption{Decision Tree Classifier Results}
  \label{res:dtree}
  \Description{Precision, recall, f1 scores and accuracy of Decision Tree Classifier.}
\end{figure}

\begin{figure}[h]
  \centering
  \includegraphics[width=0.5\linewidth]{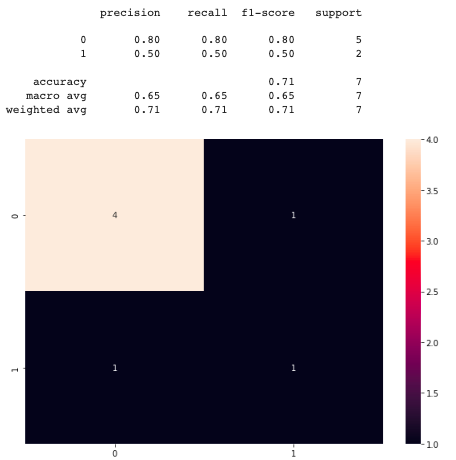}
  \caption{Random Forest Classifier Results}
  \label{res:rand_forest}
  \Description{Precision, recall, f1 scores and accuracy of Random Forest Classifier.}
\end{figure}

We calculated the mean of the well-being scores of all rooms, and labeled the rooms with a score higher than the mean as “better-than-average”, and the rooms with a score lower than the mean as “worse than average”. After labeling the rooms into two categories, we trained a set of binary classifiers on these categories. We used KNN Classification, Decision Tree Classification and Random Forest Classification respectively. Respectively, the results, including precision, recall and f1 score for the two labels, and the model accuracy shown in Figure \ref{res:knn}, \ref{res:dtree} and \ref{res:rand_forest} respectively.

Note that Label 1 refers to the better-than-average category and label 0 refers to the worse-than-average category.

As a reminder, selecting a subset of features based on their Pearson Correlation Coefficients is an effort made to shrink the model's feature dimension. It is worth noting that the Pearson Correlation Coefficients calculated in primary steps are only showing the mathematical relevancy of features and the well-bring score ground-truths. When it comes to trained models, feature relevancy in decision models might reveal different results of relevancy. Given more data entries, it would be a reasonable next step to look at the feature relevancy or "weight" in the trained models to infer further arguments on the significance of different environmental inputs. More promising future work directions are discussed in Section 7.


\section{Discussion \& Future Work}

Given that a prediction result better than random guess is made from the classification experiments based on the very limited amount of data we collected so far, we find this line of work carrying great potentials. We reflect on the limitations of the current work's data collection, feature engineering and model training methods, and propose potential rewarding directions of future work.

\subsection{Limitations}

Due to our limitation on resources, there is a natural shortage of participants and their effort in participating in this study. A total number of 22 samples does not provide a solid machine learning model to convey our research ideas. However, the result does offer a somewhat reliable conclusion about the potential of our approach to predict well-being using environmental factors and Feng Shui principles. 

This research is also limited by the length of participants staying in that specific space. Living is a long-term, complex interaction with the environment consisting of different kinds of activities. Answering survey questions for less than half an hour might not provide enough interaction for a time that is long enough to capture all influence of the environment on people. 

More importantly, the time frame this study was conducted was within the weeks towards the final at Cornell Tech. The participant body was also limited to only Cornell Tech graduate students. This type of adjacency selection might create a certain time effect that participants might have reported higher stress levels and lower attention towards the environment due to their finals. 

Another participant factor that could limit the external validity of this research is the belief in Feng Shui. As previous researchers stated, belief might be able to explain a lot of subjective phenomena. Belief can be a factor that mediates the correlation between Feng Shui principles and perceived well-being. This has also been addressed in the survey by asking participants to report their beliefs in Feng Shui, which could be potentially controlled in the data analysis in the future.

Lastly, the measurement of well-being in this research is limited to only subjective self-reported data. More factors could be considered to validate these findings from all perspectives, especially physiological feedback that people cannot control easily. 

\subsection{Future Work}

First of all, more participants can be run for a longer time span in the future studies. This can best mimic the real living experience in the specific space, also to capture long-term changes in the environment at different time periods of a day. Factors like sunlight might change the environment’s temperature and light intensity. This could also be connected with local UV ratio to provide a better feature on how much direct sunlight a room gets over the day/year. More participants could also support a more reliable and stable machine learning model in the future.

Secondly, instead of extracting one data entry from each room, we can instead cumulate a large set of data per room from all the raw entry data collected during the survey. With a large dataset accumulated as such, we will be able to utilize suitable Deep Learning models for potentially finer predictions. 

The current effort lies in a classification model to predict either “good Feng Shui” or “bad Feng Shui”. In future anticipations, a regression model will be more suitable for this type of prediction due to the fact that the well-being score can be much closer to regression data as labels, and possible future factors measured using biofeedback. 

A mediating effect of belief can also be included in the prediction model to determine its role in the prediction relationship. Possible causation can be reasoned from this type of analysis in the future to locate the real ability of Feng Shui in predicting people’s everyday living well-being.

\section{CONCLUSION}

In conclusion, this study explores the potential of adopting traditional Feng Shui principles in predicting people’s well-being in a certain environment through scientific approaches and quantified measures. The result suggests that it has a promising potential in building up to a more accurate model, and it is possible to predict people’s wellness using environmental factors like air quality, floor plan, light intensity, and temperature. In future implementations, more sensors to evaluate the environment and more measures towards subjective well-being using a long-term longitudinal study could lead to a modern technological product in predicting everyday Feng Shui at a household level.

\bibliographystyle{ACM-Reference-Format}
\bibliography{ref}


\begin{thebibliography}{8}


\ifx \showCODEN    \undefined \def \showCODEN     #1{\unskip}     \fi
\ifx \showDOI      \undefined \def \showDOI       #1{#1}\fi
\ifx \showISBNx    \undefined \def \showISBNx     #1{\unskip}     \fi
\ifx \showISBNxiii \undefined \def \showISBNxiii  #1{\unskip}     \fi
\ifx \showISSN     \undefined \def \showISSN      #1{\unskip}     \fi
\ifx \showLCCN     \undefined \def \showLCCN      #1{\unskip}     \fi
\ifx \shownote     \undefined \def \shownote      #1{#1}          \fi
\ifx \showarticletitle \undefined \def \showarticletitle #1{#1}   \fi
\ifx \showURL      \undefined \def \showURL       {\relax}        \fi
\providecommand\bibfield[2]{#2}
\providecommand\bibinfo[2]{#2}
\providecommand\natexlab[1]{#1}
\providecommand\showeprint[2][]{arXiv:#2}

\bibitem[Barrington-Leigh et~al\mbox{.}(2019)]%
        {Barrington19}
\bibfield{author}{\bibinfo{person}{C. Barrington-Leigh}, \bibinfo{person}{J.
  Baumgartner}, \bibinfo{person}{E. Carter}, \bibinfo{person}{B.~E. Robinson},
  \bibinfo{person}{S. Tao}, {and} \bibinfo{person}{Y. Zhang}.}
  \bibinfo{year}{2019}\natexlab{}.
\newblock \showarticletitle{An evaluation of air quality, home heating and
  well-being under Beijing’s programme to eliminate household coal use.}
\newblock \bibinfo{journal}{\emph{Nature Energy}} \bibinfo{volume}{4},
  \bibinfo{number}{5} (\bibinfo{year}{2019}), \bibinfo{pages}{416--423}.
\newblock
\urldef\tempurl%
\url{https://doi.org/10.1038/s41560-019-0386-2}
\showDOI{\tempurl}


\bibitem[Bonaiuto et~al\mbox{.}(2010)]%
        {Bonaiuto10}
\bibfield{author}{\bibinfo{person}{M. Bonaiuto}, \bibinfo{person}{E. Bilotta},
  {and} \bibinfo{person}{A. Stolfa}.} \bibinfo{year}{2010}\natexlab{}.
\newblock \showarticletitle{'FENG SHUI' AND ENVIRONMENTAL PSYCHOLOGY: A
  CRITICAL COMPARISON.}
\newblock \bibinfo{journal}{\emph{Journal of Architectural and Planning
  Research}} \bibinfo{volume}{27}, \bibinfo{number}{1} (\bibinfo{year}{2010}),
  \bibinfo{pages}{23--34}.
\newblock
\urldef\tempurl%
\url{http://www.jstor.org/stable/43030890}
\showURL{%
\tempurl}


\bibitem[Boschen and Oei(2007)]%
        {Boschen07}
\bibfield{author}{\bibinfo{person}{M.J. Boschen} {and}
  \bibinfo{person}{T.~P.~S. Oei}.} \bibinfo{year}{2007}\natexlab{}.
\newblock \showarticletitle{Discriminant validity of the MASQ in a clinical
  sample.}
\newblock \bibinfo{journal}{\emph{Psychiatry Research}} \bibinfo{volume}{150},
  \bibinfo{number}{2} (\bibinfo{year}{2007}), \bibinfo{pages}{163--171}.
\newblock
\urldef\tempurl%
\url{https://doi.org/10.1016/j.psychres.2006.03.008}
\showDOI{\tempurl}


\bibitem[Gidlöf-Gunnarsson and Öhrström(2007)]%
        {Gidlof07}
\bibfield{author}{\bibinfo{person}{A. Gidlöf-Gunnarsson} {and}
  \bibinfo{person}{E. Öhrström}.} \bibinfo{year}{2007}\natexlab{}.
\newblock \showarticletitle{Noise and well-being in urban residential
  environments: The potential role of perceived availability to nearby green
  areas.}
\newblock \bibinfo{journal}{\emph{Landscape and Urban Planning}}
  \bibinfo{volume}{83}, \bibinfo{number}{2} (\bibinfo{year}{2007}),
  \bibinfo{pages}{115--126}.
\newblock
\urldef\tempurl%
\url{https://doi.org/10.1016/j.landurbplan.2007.03.003}
\showDOI{\tempurl}


\bibitem[Mak and Ng(2005)]%
        {Mak05}
\bibfield{author}{\bibinfo{person}{M.~Y. Mak} {and} \bibinfo{person}{S.~Thomas
  Ng}.} \bibinfo{year}{2005}\natexlab{}.
\newblock \showarticletitle{The art and science of Feng Shui—A study on
  architects’ perception.}
\newblock \bibinfo{journal}{\emph{Building and Environment}}
  \bibinfo{volume}{40}, \bibinfo{number}{3} (\bibinfo{year}{2005}),
  \bibinfo{pages}{427--434}.
\newblock
\urldef\tempurl%
\url{https://doi.org/10.1016/j.buildenv.2004.07.016}
\showDOI{\tempurl}


\bibitem[Noelke et~al\mbox{.}(2016)]%
        {Noelke16}
\bibfield{author}{\bibinfo{person}{C. Noelke}, \bibinfo{person}{M. McGovern},
  \bibinfo{person}{D.~J. Corsi}, \bibinfo{person}{M.~P. Jimenez},
  \bibinfo{person}{A. Stern}, \bibinfo{person}{I.~S. Wing}, {and}
  \bibinfo{person}{L. Berkman}.} \bibinfo{year}{2016}\natexlab{}.
\newblock \showarticletitle{Increasing ambient temperature reduces emotional
  well-being.}
\newblock \bibinfo{journal}{\emph{Environmental Research}}
  \bibinfo{volume}{151} (\bibinfo{year}{2016}), \bibinfo{pages}{124--129}.
\newblock
\urldef\tempurl%
\url{https://doi.org/10.1016/j.envres.2016.06.045}
\showDOI{\tempurl}


\bibitem[Righi et~al\mbox{.}(2002)]%
        {Righi02}
\bibfield{author}{\bibinfo{person}{E. Righi}, \bibinfo{person}{G. Aggazzotti},
  \bibinfo{person}{G. Fantuzzi}, \bibinfo{person}{V. Ciccarese}, {and}
  \bibinfo{person}{G. Predieri}.} \bibinfo{year}{2002}\natexlab{}.
\newblock \showarticletitle{Air quality and well-being perception in subjects
  attending university libraries in Modena (Italy).}
\newblock \bibinfo{journal}{\emph{Science of The Total Environment}}
  \bibinfo{volume}{286}, \bibinfo{number}{1} (\bibinfo{year}{2002}),
  \bibinfo{pages}{41--50}.
\newblock
\urldef\tempurl%
\url{https://doi.org/10.1016/S0048-9697(01)00960-3}
\showDOI{\tempurl}


\bibitem[Wang et~al\mbox{.}(2014)]%
        {Wang14}
\bibfield{author}{\bibinfo{person}{Y. Wang}, \bibinfo{person}{X. Huang}, {and}
  \bibinfo{person}{S. Ouyang}.} \bibinfo{year}{2014}\natexlab{}.
\newblock \bibinfo{booktitle}{\emph{Automatic Feng Shui evaluation method and
  evaluation system used for residential house type.}}
\newblock
\urldef\tempurl%
\url{https://patents.google.com/patent/CN103679618A/zh}
\showURL{%
Retrieved May 17, 2022 from \tempurl}


\end{thebibliography}

\end{document}